\documentclass[journal]{IEEEtran}
\usepackage{cite}

\ifCLASSINFOpdf
\else
\fi

\usepackage[export]{adjustbox}

\usepackage{amssymb}

\usepackage{lineno}
\usepackage{soul,color}

\usepackage[dvipsnames]{xcolor}
\usepackage{algorithm, algpseudocode}

    
\usepackage{multirow, hhline}
\usepackage{mathtools, mathdots}
\usepackage{hyperref}
\hypersetup{
    colorlinks=true,
    linkcolor=blue,
    filecolor=magenta,      
    urlcolor=blue,
}
\usepackage{caption, subcaption} 
\usepackage{textcomp}
\usepackage{gensymb}
\usepackage{diagbox}
\usepackage{pifont}
\newcommand{\cmark}{\ding{51}}%
\newcommand{\xmark}{\ding{55}}%

\usepackage{rotating} 
\usepackage{lipsum} 

\DeclareMathOperator*{\argmax}{arg\,max}
\DeclareMathOperator*{\argmin}{arg\,min}

\usepackage{array}
\newcolumntype{C}[1]{>{\centering\let\newline\\\arraybackslash\hspace{0pt}}m{#1}}
\newcolumntype{R}[1]{>{\flushright\let\newline\\\arraybackslash\hspace{0pt}}m{#1}}

\allowdisplaybreaks

\makeatletter
\renewcommand{\@IEEEsectpunct}{}
\makeatother

\newcommand{\rmk}[1]{{\bf #1}}

\newcommand{\revised}[1]{{{#1}}}

\usepackage{fancyhdr}
\fancypagestyle{plain}{
  \fancyhf{}
  \fancyhead[C]{Conference on \LaTeX}     
  \fancyfoot[L]{This is a notice}

}
\usepackage{eso-pic}

\begin{document}
\AddToShipoutPictureBG*{%
  \AtPageUpperLeft{%
    \setlength\unitlength{1in}%
    \hspace*{\dimexpr0.5\paperwidth\relax}
    \makebox(0,-0.75)[c]{\parbox{0.8\textwidth}{\centering This paper has been accepted for publication in the IEEE Transactions on Robotics.\\
    Please cite as: Placed, J. A., Strader, J., Carrillo, H., Atanasov, N., Indelman, V., Carlone, L., \& Castellanos, J. A. (2023). A survey on active simultaneous localization and mapping: State of the art and new frontiers. \textit{IEEE Transactions on Robotics (T-RO)}.}}%
}}

%
\title{A Survey on Active Simultaneous Localization and
Mapping: State of the Art and New Frontiers}


\author{
    Julio~A.~Placed*, Jared~Strader, Henry~Carrillo, \\ Nikolay~Atanasov, Vadim~Indelman, Luca~Carlone, Jos\'e~A.~Castellanos  
    \thanks{This work was partially supported by MINECO project PID2019‐108398GB‐I00, 
     DGA\_FSE T45\_20R, ARL DCIST W911NF-17-2-0181 and Israel Science Foundation (ISF), grant No. 371/20.}
    \thanks{*Corresponding author.}
    \thanks{Julio A. Placed and Jos\'e A. Castellanos are with the Instituto de Investigaci\'on en Ingenier\'ia de Arag\'on (I3A), Universidad de Zaragoza, C/Mar\'ia de Luna 1, 50018, Zaragoza, Spain (e-mail: \{jplaced, jacaste\}@unizar.es).}
    \thanks{Jared Strader and Luca Carlone are with the Laboratory for Information \& Decision Systems, Massachusetts Institute of Technology, Cambridge, MA 02139, USA (e-mail: \{jstrader, lcarlone\}@mit.edu).}
    \thanks{Henry Carrillo is with Genius Sports, One Plaza, Avenida El Poblado \#5A -- 113, North Tower, 2nd \& 10th Flr, Medellín, AN, COL (e-mail: henry.carrillo@geniussports.com).}
    \thanks{Nikolay Atanasov is with the Department of Electrical and Computer Engineering, UC San Diego, 9500 Gilman Drive, La Jolla, CA 92093-0411, USA (e-mail: natanasov@ucsd.edu).}
    \thanks{Vadim Indelman is with the Department of Aerospace Engineering, Technion -- Israel Institute of Technology, Haifa 32000, Israel. (e-mail: vadim.indelman@technion.ac.il).}
}


\maketitle

\begin{abstract}
Active Simultaneous Localization and Mapping (SLAM) is the problem of planning and controlling the motion of a robot to build the most accurate and complete model of the surrounding environment. 
Since the first foundational work in active perception appeared, more than three decades ago, this field has received increasing attention across different scientific communities. This has brought about many different approaches and formulations, and makes a review of the current trends necessary and extremely valuable for both new and experienced researchers. 
In this work, we survey the state-of-the-art in active SLAM and take an in-depth look at the open challenges that still require attention to meet the needs of modern applications. 
After providing a historical perspective, we present a unified problem formulation and review the well-established modular solution scheme, which decouples the problem into three stages that identify, select, and execute potential navigation actions. 
We then analyze alternative approaches, including belief-space planning and deep reinforcement learning techniques, and review related work on multi-robot coordination.
The manuscript concludes with a discussion of new research directions, addressing reproducible research, active spatial perception, and practical applications, among other topics.
\end{abstract}


\IEEEpeerreviewmaketitle

\section{Introduction} \label{S:1}

\IEEEPARstart{A}{utonomous} operation in robotics applications requires robots to have access to a consistent model of the surrounding environment, in order to support safe planning and decision making. 
Towards this goal, 
a robot must have the ability to create a map of the environment, localize itself on it, and control its own motion. Active SLAM refers to the joint resolution of these three core problems in mobile robotics, with the ultimate goal of creating the most accurate and complete model of an unknown environment. 
Active SLAM can be seen as a
decision-making process in which the robot has to choose its own future control actions, balancing between exploring new areas and exploiting those already seen to improve the accuracy of the resulting map model.

During the last decades, active SLAM has received increasing attention\footnote{The number of publications on active SLAM has grown from 53 in 2010 to over 660 in 2022 (a twelve-fold increase). The number becomes almost 5500 if we extend the search to include belief-space planning, active exploration, and simultaneous planning, localization and mapping.
Source: dimensions.ai.}
and has been studied in different forms across multiple communities, with the ambition of deploying autonomous agents in real-world applications (e.g.,~search and rescue in hazardous environments, underground or planetary exploration). This divergence has broadened the scope of the problem and provided a wider context, yielding numerous approaches based on different concepts and theories that have made the field flourish; but it also created a disconnect between research lines that could mutually benefit from each other. 
With this survey, we seek to fill this gap by providing a general problem statement and a unified review of related works.

Currently, active SLAM is at a decisive point, 
driven by novel opportunities in spatial perception and artificial intelligence (AI). 
These include, for instance, the application of breakthroughs in neural networks to prediction beyond line-of-sight, reasoning over novel environment representations, or leveraging new SLAM techniques to process dynamic and deformable scenes. Throughout this paper, we give a fresher picture of active SLAM that goes beyond the classical
---but still mainstream--- entropy computation over discretized grids. Besides, we identify the open challenges that need to be addressed for active SLAM to have an impact on real applications, shaping future lines of research, and describing how they can nourish from the cross-fertilization between research fields. Among those challenges, we emphasize the urgent need for benchmarks and reproducible research.

\subsection{Historical Perspective}
Ever since the first mobile robots were built in the late 1940s, the ambition that they could perform autonomous tasks has been one of the major focuses of robotics research. To operate autonomously, a robot needs to  form a model of the surrounding environment ---including localization and mapping--- and perform safe navigation~\cite{siegwart11}.
While the former involves estimating the position of the robot and creating a symbolic representation of the environment, the latter refers to planning and controlling the movements of the robot to safely achieve a goal location. 
Localization, mapping, and planning have been often investigated 
in combination, resulting in multiple research areas such as SLAM, active localization, active mapping, and active SLAM.

Localization and mapping were treated deterministically and solved independently until probabilistic approaches went mainstream in the 1990s, when researchers realized that both tasks were correlated and dependent of one another. SLAM refers, thereby, to the problem of incrementally building the map of an environment while at the same time locating the robot within it~\cite{thrun05}. This problem has attracted significant attention from the robotics community in the last decades;  
see~\cite{durrant06, grisetti10, cadena16} and the references therein.

SLAM, however, is a passive method and is not concerned with guiding the navigation process. In contrast, \emph{active} approaches do consider the navigation aspects of the problem. 
Bajcsy~\cite{bajcsy85}, Cowan and Kovesi~\cite{cowan88}, and Aloimonos \textit{et al.}~\cite{aloimonos88} were the first to study and analyze the problem of active perception (also referred to as active information acquisition~\cite{atanasov14}) in the late nineties. Bajcsy~\cite{bajcsy98} would later formally define it as the problem of actively acquiring data in order to achieve a certain goal, necessarily involving a decision-making process. 
For the cases in which the objective is to improve localization, mapping, or both, the problems are respectively referred to as active localization, active mapping, and active SLAM.

\emph{Active mapping} was the first problem to be addressed, dating back to the work of Connolly~\cite{connolly85} in 1985. Better known since then as the \emph{next best view} problem, active mapping tackles the search of the optimal movements to create the best possible representation of an environment. Subsequent examples date to the 1990s~\cite{maver93, whaite97, pito99}, always under the assumption of perfectly known sensor localization. This problem has been primarily addressed in the computer vision community to reconstruct objects and scenes from multiple viewpoints, since the nature of the projective geometry for monocular cameras, occlusions, and limited field of view often make impossible to do it from just one viewpoint; see~\cite{zeng20b} and the references therein. 

In a similar vein, \emph{active localization} aims to improve the estimation of the robot's pose by determining how it should move, assuming the map of the environment is known. First relevant works can be traced back to 1998, when Fox \textit{et al.}~\cite{fox98} and Borgi and Caglioti~\cite{borghi98} formulated it as the problem of determining the robot motion so as to minimize its future expected (i.e.,~\textit{a posteriori}) uncertainty. In particular, it is in~\cite{fox98} where the foundations of the current workflow were laid: (i) goal identification, (ii) utility computation, and (iii) action selection (we will extensively review these stages later in this survey). Other relevant subsequent work can be found in~\cite{jensfelt01, mostegel14, gottipati19, xie20}, but also in the related literature of perception-aware planning~\cite{strader20} and planning under uncertainty~\cite{roy99}.

Finally, \emph{active SLAM} unifies the previous problems, and allows a robot to operate autonomously in an initially unknown environment. It refers to the application of active perception to SLAM and can be defined as the problem of controlling a robot which is performing SLAM in order to reduce the uncertainty of its localization and the map representation~\cite{carrillo12}.
Historically, active SLAM has been referred to with different terminology, which has significantly hindered knowledge sharing and dissemination within the robotics community. Relevant seminal works can be found under the names of active exploration~\cite{thrun91}, adaptive exploration~\cite{feder99, bourgault02}, integrated exploration~\cite{makarenko02, stachniss04}, autonomous SLAM~\cite{newman03}, simultaneous planning, localization and mapping~\cite{stachniss09}, belief-space planning (BSP)~\cite{platt10}, or simply robotic exploration~\cite{stachniss03, sim05}. It was not until 2002 ---when Davison and Murray~\cite{davison02} coined the term active SLAM--- that the robotics community started adopting this nomenclature. 
Thrun and M\"{o}ller~\cite{thrun91} demonstrate that in order to solve robotic exploration, agents have to switch between two opposite principles depending on the expected costs and gains: 
exploring new areas and revisiting those already seen, 
i.e.,~the so-called \emph{exploration-exploitation dilemma}. The first approach in which a robot chooses actions that maximize the knowledge of the two variables of interest is attributed to Feder \textit{et al.}~\cite{feder99}, who also separate the procedure in three major stages as in~\cite{fox98}. Table~\ref{tab:works} contains a subset of relevant works that have followed~\cite{feder99}. 
This table differentiates the main aspects of each approach, including the type of sensors, the state representation, 
 and the theoretical foundations. 

\subsection{About Previous Surveys}

Only two works have previously addressed the problem of surveying active SLAM research. The first of them, published in 2016, is a section of a more general survey on SLAM carried out by Cadena \textit{et al.}~\cite{cadena16}. The other, by Lluvia \textit{et al.}~\cite{lluvia21}, conducts a more extensive survey on ``Active Mapping and Robot Exploration''. Table~\ref{tab:surveys} summarizes the topics they address, along with those covered in the present survey.

Cadena \textit{et al.}~\cite{cadena16} describe both the history and the main aspects of the problem, and identify three open challenges: the decision of when to stop performing active SLAM, the problem of accurately predicting the effect of future actions, and the lack of mathematical guarantees of optimality. However, the brevity of the active SLAM section prevented delving 
into a detailed discussion of the most relevant works or providing a more unified mathematical formulation of the problem. Moreover, since~\cite{cadena16} was published, many relevant contributions have been proposed and new open problems have arisen. For instance, progress has been made on the way uncertainties of the robot location and the map are represented and quantified. Furthermore, recent work has also opened new research endeavors, including deep learning (DL).

Lluvia \textit{et al.}~\cite{lluvia21} also provide a thorough historical review and relate the different communities that have been trying to solve this problem under different nomenclatures. Similar to~\cite{cadena16}, they do not attempt to present a unified mathematical formulation of active SLAM nor do they cover utility computation, a field which has been mostly overlooked in the literature.
They delve, nevertheless, into the optimization of vantage points and the trajectories to reach them, a new problem that has attracted significant attention from the control community and has seen many contributions in recent years. 
In~\cite{lluvia21}, the authors present a comparison between representative works in active SLAM, although with a limited scope. Contrarily to~\cite{lluvia21}, we present a more complete analysis and a broader set of open challenges, which extends the ones identified in~\cite{cadena16}.

\renewcommand{\arraystretch}{1.1}
\begin{sidewaystable*}
    \footnotesize
    \centering
    \begin{tabular}{R{9.5em}|| C{8em} | C{6.8em} | C{7em} | C{5.4em} | C{7em} | C{9em} | C{6em} | C{5em} | C{3.8em}}
        \textbf{Reference} & \textbf{SLAM Approach} & \textbf{Sensors} & \textbf{Environment Representation} & \textbf{Formulation} & \textbf{\revised{Candidate} Goal Locations} & \textbf{Utility Function} & \textbf{Validation Environment} & \textbf{Stopping Criterion} & \revised{\textbf{Publicly available}} \\ \hline \hline
        Feder \textit{et al.}~\cite{feder99} & EKF & Sonar & Landmark map & \revised{Modular} & Local vicinity & $D\text{-}opt$ & Sim. \& real & - & \revised{\xmark}\\ \hline
        Bourgault \textit{et al.}~\cite{bourgault02} & EKF & Lidar & OG map & \revised{Modular} & Local vicinity & MI & Real & - & \revised{\xmark}\\ \hline
        Stachniss \textit{et al.}~\cite{stachniss04} & FastSLAM~\cite{montemerlo02} & Lidar & OG map & \revised{Modular} & Frontiers \& re-visiting & Particle's volume \& distance & Sim. \& real & Particle's volume & \revised{\xmark}\\ \hline
        Stachniss \textit{et al.}~\cite{stachniss05} & RBPF & Lidar & OG map & \revised{Modular} & Frontiers \& re-visiting & MI \& distance & Sim. \& real & - & \revised{\xmark}\\ \hline
        Leung \textit{et al.}~\cite{leung06} & EKF & Lidar & Landmark map & MPC & Unknown space \& re-visiting & $T\text{-}opt$ & Simulation & - & \revised{\xmark}\\ \hline
        Valencia \textit{et al.}~\cite{valencia12} & Pose SLAM~\cite{ila09} & Lidar & OG map & \revised{Modular} & Frontiers \& revisiting & Entropy & Simulation & - & \revised{\xmark}\\ \hline
        Carlone \textit{et al.}~\cite{carlone14jirs} & RBPF & Lidar & OG map & \revised{Modular} & Frontiers \& re-visiting & KLD & Simulation & - & \revised{\xmark}\\ \hline  
        Indelman \textit{et al.}~\cite{indelman15} & GTSAM & Camera & Landmark map & BSP & - & Robot's $T\text{-}opt$ \& distance & Simulation & - & \revised{\xmark}\\ \hline
        Zhu \textit{et al.}~\cite{zhu15} & RGBDSLAM~\cite{endres13} & RGB-D & Octomap & \revised{Modular} & Frontiers & Coverage \& distance & Simulation & - & \revised{\xmark}\\ \hline
        Bircher \textit{et al.}~\cite{bircher16} & ROVIO~\cite{bloesch15} \& mapping & Stereo \& IMU & Octomap & MPC & RRT paths & Coverage \& distance & Sim. \& real & Coverage & \revised{\cmark} \\ \hline
        Papachristos \textit{et al.}~\cite{papachristos17} & ROVIO~\cite{bloesch15} \& dense mapping & Stereo \& IMU & Octomap & MPC & RRT paths & $D\text{-}opt$ & Sim. \& real & Min. utility & \revised{\cmark}\\ \hline
        Umari \textit{et al.}~\cite{umari17} & Gmapping~\cite{grisetti07} & Lidar & OG map & \revised{Modular} & Frontiers & Map's MI \& distance & Sim. \& real & -  & \revised{\cmark}\\ \hline
        Carrillo \textit{et al.}~\cite{carrillo18} & ICP \& iSAM~\cite{kaess08} & Lidar & OG map & \revised{Modular} & Frontiers & Shannon-R\'enyi entropy & Sim. \& real & Time &\revised{\cmark} \\ \hline
        Jadidi \textit{et al.}~\cite{jadidi18} & Pose SLAM~\cite{ila09} & Lidar & COM & \revised{Modular} & Frontiers & MI \& distance & Simulation & Coverage & \revised{\cmark} \\ \hline
        Palomeras \textit{et al.}~\cite{palomeras19b} & ICP \& g2o~\cite{grisetti11} & Lidar & Octomap & \revised{Modular} & Random & Coverage & Sim. \& real & Coverage & \revised{\xmark}\\ \hline
        Chaplot \textit{et al.}~\cite{chaplot20} & Neural Networks & RGB & OG map & DRL & Local vicinity & Coverage & Sim. \& real & - & \revised{\cmark}\\ \hline
        Niroui \textit{et al.}~\cite{niroui19} & Gmapping~\cite{grisetti07} & Lidar & OG map & DRL & Frontiers & Map's MI \& distance & Sim. \& real & - & \revised{\xmark}\\ \hline
        \revised{Chen \textit{et al.}~\cite{chen20c}} & \revised{GTSAM} & \revised{Range} & \revised{Virtual landmark map} & \revised{DRL} & \revised{Frontiers} & \revised{Virtual landmark's $T\text{-}opt$} & \revised{Simulation} & \revised{-} &\revised{\cmark}\\ \hline
        \revised{Li \textit{et al.}~\cite{li20}} & \revised{Karto~\cite{konolige10} \& g2o~\cite{grisetti11}} & \revised{Lidar} & \revised{OG map} & \revised{DRL} & \revised{Sampled from the OG map} & \revised{Map's MI \& distance} & \revised{Sim. \& real} & \revised{Coverage} &\revised{\xmark}\\ \hline
        Suresh \textit{et al.}~\cite{suresh20} & ICP \& iSAM~\cite{kaess08} & Sonar & Octomap & MPC & RRT paths \& re-visiting & Robot's $D\text{-}opt$ \& coverage & Sim. \& real & - & \revised{\xmark}\\ \hline
        Chen \textit{et al.}~\cite{chen20b} & Linear SLAM~\cite{zhao19} & Camera & Landmark map & MPC & Local vicinity & Graph's $D\text{-}opt$ & Sim. \& real & Coverage & \revised{\xmark}\\ \hline
        \revised{Batinovic \textit{et al.}~\cite{batinovic21}} & \revised{Cartographer~\cite{hess16}} & \revised{Lidar \& IMU} & \revised{Octomap} & \revised{Modular} & \revised{Frontiers} & \revised{Map's MI \& distance} & \revised{Sim. \& real} & \revised{Coverage} &\revised{\cmark}\\ \hline
        \revised{Placed \textit{et al.}~\cite{placed22b}} & \revised{ORB-SLAM2~\cite{mur17}} & \revised{Lidar \& RGB-D} & \revised{Octomap} & \revised{Modular} & \revised{Frontiers} & \revised{Graph's $D\text{-}opt$} & \revised{Simulation} & \revised{Time} &\revised{\cmark}\\ \hline
        \revised{Bonetto \textit{et al.}~\cite{bonetto22}} & \revised{RTAB-Map~\cite{labbe19}} & \revised{Lidar, RGB-D \& IMU} & \revised{Octomap} & \revised{Modular \& MPC} & \revised{Frontiers} & \revised{Map's MI, distance \& visual features} & \revised{Sim. \& real} & \revised{Time} & \revised{\cmark}\\
    \end{tabular}
    \caption{A comparison between representative active SLAM approaches, ordered chronologically.}
    \label{tab:works}
\end{sidewaystable*}

\renewcommand{\arraystretch}{1.2}
\begin{table}[t!]
    \centering
    \begin{tabular}{l l||C{3.6em}|C{3.9em}|c}
        \multicolumn{2}{c||}{\textbf{Topic}}& \textbf{Cadena \textit{et al.}}~\cite{cadena16} & \textbf{Lluvia \textit{et al.}}~\cite{lluvia21} & \textbf{\textit{Ours}} \\ \hline \hline
        \multirow{2}{*}{Introduction}  & Historical review & Briefly & Yes & Yes \\ 
        & Problem formulation & No & No & Yes \\ \hline
        \multirow{5}{*}{\shortstack[l]{\revised{Modular}\\ scheme}} & Env. representation & Yes & Yes & Yes \\
        & Goal identification & Briefly & Yes & Yes \\ 
        & Information Theory & Briefly & No & Yes \\
        & TOED & Briefly & No & Yes \\
        & Graph Theory & No & No & Yes \\ \hline
        \multirow{3}{*}{\shortstack[l]{Alternative\\ approaches}} 
        & Continuous domain & No & Yes & Yes \\
        & Deep Learning & No & Briefly & Yes \\
        & Multi-robot & No & No & Yes \\ \hline
        \multirow{7}{*}{\shortstack[l]{Open\\ problems}} & State prediction & Yes & Yes & Yes \\
        & Stopping criteria & Yes & Briefly & Yes \\
        & Novel representations & No & Briefly & Yes \\
        & Data association & No & No & Yes \\
        & Complex environments & No & No & Yes \\
        & Reproducible research & No & No & Yes \\
        & Practical applications & No & No & Yes \\
    \end{tabular}
    \caption{Comparison between the topics and open challenges addressed in previous surveys and the current one.}
    \label{tab:surveys}
\end{table}
\subsection{Paper Structure} 
The remainder of this manuscript is organized as follows. Section~\ref{S:2} provides a unified problem formulation for active SLAM and describes the three subproblems (or \emph{stages}) it has traditionally been divided into. 
Sections~\ref{S:3} to~\ref{S:5} cover those three stages separately. In particular, Section~\ref{S:3} deals with the identification of vantage points, Section~\ref{S:4} with utility computation, and Section~\ref{S:5} with selection and execution of the optimal action. 
Sections~\ref{S:6} and~\ref{S:7} consider, on the other hand, alternative continuous-state optimization and DL methods.
 Section~\ref{S:8} is devoted to multi-robot active SLAM. 
 Section~\ref{S:9} outlines the open research questions in active SLAM. 
 Finally, Section~\ref{S:10} concludes the manuscript. 

\section{The Active SLAM Problem} \label{S:2}

\subsection{Problem Formulation} \label{SS:2a}

Active SLAM 
can be framed within the wider mathematical framework of partially observable Markov decision processes (POMDPs), after some particularization. 
POMDPs model decision-making problems under both action and observation uncertainties and can be formally defined as the 7-tuple $(\mathcal{S},\mathcal{A},\mathcal{Z},\xi_s,\xi_z,r,\gamma)$. In particular, a POMDP consists of the agent's state space $\mathcal{S}$, a set of actions $\mathcal{A}$, a transition function between states $\xi_s:\mathcal{S}\times\mathcal{A}\mapsto\Pi(\mathcal{S})$ where $\Pi(\mathcal{S})$ is the space of probability density functions (pdfs) over $\mathcal{S}$, an observation space $\mathcal{Z}$, the conditional likelihood of making any of those observations $\xi_z:\mathcal{S}\mapsto\Pi(\mathcal{Z})$, where $\Pi(\mathcal{Z})$ is the space of pdfs over $\mathcal{Z}$, a reward scalar mapping $r:\mathcal{S}\times\mathcal{A}\to\mathbb{R}$, and the discount factor $\gamma\in(0,1)\in\mathbb{R}$ which allows to work with finite rewards even when planning over infinite time horizons.

Contrary to the fully observable case, agents in a POMDP cannot reliably determine their own true state, $\boldsymbol{s}$. Instead, they maintain an internal \textit{belief} or \textit{information state}, $b_t(\boldsymbol{s}_t)$, which represents the posterior probability over states at time $t$, given the available data collected up to that time~\cite{kaelbling98, thrun05, sigaud13}:
\begin{equation}
    b_t(\boldsymbol{s}_t)\triangleq p(\boldsymbol{s}_t|\underbrace{\boldsymbol{z}_{1:t},\boldsymbol{a}_{1:t-1}}_{\text{history, }\boldsymbol{h}}) \, ,\label{eq:belief}
\end{equation}
where $\boldsymbol{z}_{1:t}$ is the set of all available observations and  $\boldsymbol{a}_{1:t-1}$ the set of past control actions (both collectively referred as the \emph{history} $\boldsymbol{h}$).
%
The \emph{belief space}, $\mathcal{B}(\mathcal{S})\equiv\Pi(\mathcal{S})$, of probability density functions over the set $\mathcal{S}$ is defined as:
\begin{equation}
    \mathcal{B}(\mathcal{S}) \triangleq \{b:\mathcal{S} \mapsto \mathbb{R} \mid \int b(s) ds = 1, \ b(s) \geq 0\} \, .
\end{equation}

In order to evaluate the effect of future actions, agents must be capable of predicting posterior belief distributions, that is, the pdf over $\mathcal{S}$ after performing a certain action, $\boldsymbol{a}_t$, and taking a future observation $\boldsymbol{z}_{t+1}$:
%
\begin{align}
    b_{t+1}(\boldsymbol{s}_{t+1})&\triangleq p(\boldsymbol{s}_{t+1}|\boldsymbol{z}_{t+1}, \boldsymbol{a}_t, b_t(\boldsymbol{s}_t)) \, .
\end{align}
Since the future measurements are unknown for the agent, their expected value has to be studied instead. Consider that an agent in the state defined by $b_t(\boldsymbol{s}_t)$
executes a certain action $\boldsymbol{a}_t$, and transitions to another state with pdf $p(\boldsymbol{s}_{t+1})$. Then, the likelihood of making an observation will be given by~\cite{sigaud13}:
\begin{align}
    p(\boldsymbol{z}_{t+1}|b_t(\boldsymbol{s}_t),\boldsymbol{a}_t)=
    \int\int
    &\xi_z(\boldsymbol{s}_{t+1}) \ \xi_s(\boldsymbol{s}_t,\boldsymbol{a}_t) \nonumber\\
    & \ b_t(\boldsymbol{s}_t) 
    \ d\boldsymbol{s}_t\ d\boldsymbol{s}_{t+1}\, ,
    \label{eq:obs}
\end{align}
where $\xi_z(\boldsymbol{s}_{t+1})=p(\boldsymbol{z}_{t+1}|\boldsymbol{s}_{t+1})$ is the observation model and $\xi_s(\boldsymbol{s}_t,\boldsymbol{a}_t)=p(\boldsymbol{s}_{t+1}|\boldsymbol{s}_t,\boldsymbol{a}_t)$ the motion model.

Since the belief is a sufficient statistic, optimal policies for the original POMDP may be found by solving an equivalent continuous-space MDP over $\mathcal{B}(\mathcal{S})$~\cite{astrom65,kaelbling98}. Such MDP is defined by the 5-tuple  $(\mathcal{B},\mathcal{A},\xi_b,\rho, \gamma)$, where the transition and reward functions are $\xi_b:\mathcal{B}\times\mathcal{A}\mapsto\Pi(\mathcal{B})$ and $\rho:\mathcal{B}\times\mathcal{A}\mapsto\mathbb{R}$. To preserve consistency, this belief-dependent reward function builds on the expected rewards of the original POMDP:
\begin{equation}
    \rho(b_t,\boldsymbol{a}_t)=
    \int_\mathcal{S}
    b_t(\boldsymbol{s}_t) \ r(\boldsymbol{s}_t,\boldsymbol{a}_t) 
    \ d\boldsymbol{s}_t \, .
    \label{eq:reward}
\end{equation}

Then, the decision at time $t$ will be provided by the (control/action) policy $\pi_t$, which maps elements from the space of pdfs over $\mathcal{S}$ to the action space:
\begin{equation}
    \pi_t: \mathcal{B}(\mathcal{S}) \mapsto \mathcal{A}\, .
    \label{eq:decision_mapping}
\end{equation}
The optimal policy, $\pi^\star$, that yields the highest expected rewards for every belief state can be found via:
\begin{equation}
    \pi^\star(b) = \argmax_\pi \sum_{t=0}^{\infty}\mathbb{E}\left[ \gamma^t \rho(b_t, 
    \pi(b_t)) \right]\, ,\label{eq:optimal-policy}
\end{equation}
where expectation is taken w.r.t. $p(\boldsymbol{z}_{t+1}|b_t(\boldsymbol{s}_t),\boldsymbol{a}_t)$.
In general, computing the optimal policy for MDPs with continuous state spaces is hard and most works resort to approximate solutions or problem simplifications~\cite{kaelbling98, araya10}. 


The active SLAM problem requires, however, some variation and particularization of the above general POMDP formulation. Let us consider a robot capable of moving in an unknown environment while performing SLAM. That is, at every time step, the robot can change its own linear and angular velocities; moreover, the robot is able to process the sensor data into a map representation, $\boldsymbol{m}_t \in\mathcal{M}$, and an estimate of its own state (e.g.,~pose), 
$\boldsymbol{x}_t\in \mathcal{X}$.
Thus, the state space can be defined as the joint space $\mathcal{S}\triangleq\mathcal{X}\times\mathcal{M}$.

The evolution of both the state and the measurements in SLAM is governed by probabilistic laws~\cite{thrun05}, as~\eqref{eq:belief} and~\eqref{eq:obs} express. However, two assumptions are worth mentioning in the context of active SLAM regarding each of the equations, that further simplify its resolution. First, the robot state is commonly assumed Gaussian with a pdf $b(\boldsymbol{x})$ having mean \revised{ $\hat{\boldsymbol{x}}$ and covariance $\boldsymbol{\Sigma}_r$} (see, e.g.,~\cite{valencia12, indelman15}). \revised{Thus, the map and the robot state are usually treated independently, although some representations allow for a joint distribution (e.g.,~in sparse landmark maps or using Gaussian Processes to model dense maps~\cite{jadidi18}).} Secondly, despite less prevalent than the former, some works (e.g.,~\cite{platt10}) also assume \textit{maximum likelihood} (ML) observations, \revised{i.e.,~that executing an action in a given belief state will always produce the same, most probable observation. This allows to rewrite the expected measurements as:}
\begin{equation}\label{eq:ml_measurement}
    \boldsymbol{z}_{t+1}^{ML}=\argmax_{\boldsymbol{z}\in\mathcal{Z}} \ p(\boldsymbol{z}_{t+1}|b_t(\boldsymbol{s}_t),\boldsymbol{a}_t)\, .
\end{equation}
In addition, in active SLAM the reward typically reflects the agent's knowledge of the system (i.e.,~it involves the uncertainty in the belief rather than focusing on reaching specific states). 
These reward functions are known as \emph{utility functions} and
may be defined mathematically as the scalar mapping $ \rho: \mathcal{B}(\mathcal{S}) \times \mathcal{A} \mapsto \mathbb{R}$. This reward mapping, however, is inconsistent with both POMDPs (where the reward is dependent on $\boldsymbol{s}$ and $\boldsymbol{a}$) and belief MDPs (where the reward is restricted to the form in~\eqref{eq:reward}). To circumvent this limitation, $\rho$-POMDP~\cite{araya10} extends the POMDP formulation to allow the inclusion of beliefs' uncertainty in the objective. This enables the use of information-oriented criteria rather than control-oriented, without losing basic properties such as Markovianity.

Finally, considering a finite-horizon and ML observations, the discount factor and expectation over future measurements in~\eqref{eq:optimal-policy} can be dropped, and active SLAM can be reduced to the following optimization for open-loop planning settings:
\begin{equation}
    \boldsymbol{a}_{t:t+k}^\star = \argmax_{\boldsymbol{a}_{t:t+k}\in\revised{\mathcal{A}^k}}\ \revised{\sum_{\tau=t}^{t+k}\rho\left(b(\boldsymbol{s}_{\tau}), \boldsymbol{a}_\tau\right)}\, ,
    \label{eq:argmax_eq}
\end{equation}
where $\boldsymbol{a}_{t:t+k}^\star$ is the optimal sequence of actions to execute over the future planning horizon ($k$ lookahead steps) \revised{and $\mathcal{A}^k\triangleq\mathcal{A}\times\mathcal{A}\times...\times\mathcal{A}$ the space of sequences of actions over $k$}. 


\subsection{Decoupling Active SLAM into Three Subproblems} 

While the previous section provided a unified formulation for active SLAM, for computational convenience 
active SLAM has been traditionally decoupled into three subproblems (or \emph{stages})~\cite{fox98, feder99, makarenko02}, which will be briefly described hereafter and covered in detail in Sections~\ref{S:3} to~\ref{S:5}:
\begin{enumerate}
    \item[1)] \emph{Identification of the potential actions:} solely to reduce the computational burden, the first stage aims to determine a reduced subset of possible actions to execute.
    \item[2)] \emph{Utility computation:} the expected cost and gain of performing each candidate action has to be estimated.
    \item[3)] \emph{Action selection and execution:} finally, the last stage involves finding and executing the optimal action(s).
\end{enumerate}
The entire process should be iteratively repeated until the whole environment is accurately modeled, although in practice it is done until some stopping conditions are met.

\revised{For clarity of presentation and because many existing works do decompose active SLAM into these stages, we review each stage separately in Sections~\ref{S:3} to~\ref{S:5}.
However, this decoupling can produce suboptimal results and lead to undesired behaviors. Performing the three stages simultaneously is certainly advantageous, e.g.~when optimizing over a continuous action space, or when a control policy is optimized or learned under the umbrella of POMDPs. We review these approaches, alternative to the modular scheme, in Sections~\ref{S:6} and~\ref{S:7}.}



\section{Stage 1: Identification of Potential Actions} \label{S:3}

The first stage in \revised{modular} active SLAM approaches consists in generating the set of available actions the robot could execute (i.e.,~goals the robot can reach); 
this can be understood as a way to reduce (and discretize) the search space of potential actions.
Early works simply used random goals or required human interaction, until the concept of \emph{frontiers} was introduced by Yamauchi~\cite{yamauchi97}. This resulted in improved exploration strategies, and has consolidated as the most common approach. Nevertheless, the advent of neural networks has led to new ways of evaluating the space of potential goals. In this section, we present the most important methods to identify goal locations. Since they strongly depend on the representation of the environment estimated by the SLAM pipeline, we start by providing a brief description of the different existing representations for active SLAM.

\subsection{Representation of the Environment}

We review four different types of map representations: topological, metric, metric-semantic, and hybrid maps. 

\subsubsection{Topological maps } 
use lightweight graphs to describe information about the topology of the environment. Historically, vertices in this graph represent convex regions in the free space, while edges model connections between them. The construction of these graphs is a segmentation problem, usually done over an occupancy grid; 
see~\cite{bormann16} for a survey on these methods. Despite these maps allow leveraging graph theory, 
which provides powerful tools for planning and exploration,
they are not frequently used in active SLAM~\cite{mu16, fermin17}.

\subsubsection{Metric maps } are the most used representations to encode information about the environment in active SLAM. They can be further divided into two categories: sparse and dense maps. 
The former rely on a sparse set of interest points (or~\emph{landmarks}) to represent a scene, and have been especially used in optimal control~\cite{leung06,atanasov15,chen20b} and belief-space planning~\cite{indelman15} approaches.
Dense maps can be based on point clouds, meshes or, more typically, a discretization of the environment into cells that encode a certain metric (e.g.,~occupancy, distance to obstacles). Occupancy grid (OG) maps, first proposed in the late eighties for perception and navigation by Elfes~\cite{elfes89} and Moravec~\cite{moravec89}, assign to each cell its probability of being occupied. They have been used in numerous active SLAM frameworks, e.g.,~\cite{carrillo18, chaplot20, niroui19, placed21iros}. 
Their extension to 3D include OctoMaps~\cite{hornung13}, Supereight~\cite{vespa18} and voxel maps~\cite{muglikar20}, all of which have been also used in active SLAM~\cite{palomeras19, selin19, deng20b, dai20}. Jadidi \textit{et al.}~\cite{jadidi18} use continuous occupancy maps (COM) to leverage continuous optimization methods. There exist many other dense maps that encode more sophisticated metrics, such as those based on signed distance fields (SDF), like Voxblox~\cite{oleynikova17}. Still, they are seldom used in active SLAM~\cite{saulnier20}.

\subsubsection{Metric-semantic maps } 
go beyond geometric modeling and associate semantic information to classical metric maps. Instead of geometric features, a sparse map can capture objects, described by a semantic category, pose, and shape~\cite{bowman17,nicholson18}. Active object-level SLAM has been considered in~\cite{eidenberger10,atanasov14b}.
Examples of dense metric-semantic maps include Voxblox++~\cite{grinvald19} and Kimera~\cite{rosinol20} (which build upon an SDF), and Fusion++~\cite{mccormac18} and~\cite{zheng19} (based on voxel maps).
Despite being used in some SLAM formulations (see~\cite{cadena16,rosinol20} and the references therein), they have not yet been used in active SLAM. An exception is the work of Asgharivaskasi and Atanasov~\cite{asgharivaskasi21,asgharivaskasi22} which develops a multi-class (semantic) OctoMap and uses a closed-form lower bound on the Shannon mutual information between the map and range-category observations to select informative robot trajectories.


\subsubsection{Hybrid and hierarchical maps } combine some of the previous representations to enhance the decision-making process. Hybrid metric-topological maps have been applied to tackle either navigation~\cite{thrun96} or SLAM~\cite{tomatis03}. 
Rosinol \textit{et al.}~\cite{rosinol21} combine metric, semantic, and topological representations into a single model, a \emph{3D scene graph}. These hierarchical representations break down metric-semantic maps into interconnected high-level entities, paving the way for high-level reasoning. The use of hybrid maps in active SLAM is mostly unexplored, with~\cite{gomez19} among the few works that have integrated them.

\subsection{Detecting Goal Locations} 

The identification of all possible destinations the robot could travel to easily proves to be intractable because of the dimensions of the map and the action set~\cite{burgard05}. In practice, a finite subset of them is identified, allowing for computational tractability despite not guaranteeing global optimality~\cite{indelman15}. The simplest approach consists of randomly selecting the goal destinations~\cite{gonzalez02, tovar06}. \revised{\textbf{Random exploration}} requires low computational resources and works under the assumption that every spot in the environment has the same information associated. In 1997, Yamauchi~\cite{yamauchi97} revolutionized the field by introducing the concept of \emph{frontiers}, i.e.,~the areas that lie between known and unknown regions. Since its proposal, \revised{\textbf{frontier-based exploration}} has been the most used by far and has been tailored to different map representations.
Frontiers have been effectively identified for topological maps as nodes with no neighbors in certain directions~\cite{mu16}. 
For 2D OG maps, a plethora of geometric frontier-detection methods have been developed to circumvent the computational cost of searching the entire space~\cite{quin21}.
 Keidar and Kaminka~\cite{keidar12} propose the wavefront frontier detector (WFD) and fast frontier detector (FFD). WFD starts the search from the robot's location and restricts it to the free space; FFD performs the search after each scan is collected, following the intuition that frontiers are bound to appear in recently scanned regions. Following this idea, the same authors present the incremental WFD~\cite{keidar14}, that restricts the search to recently scanned areas. Quin \textit{et al.}~\cite{quin21} improve the performance of the previous algorithms by only evaluating a subset of the observed free space. Refer to~\cite{holz10,quin21} for further discussion. Umari and Mukhopadhyay~\cite{umari17} first present a frontier search method over a 2D OG based on rapidly-exploring random trees (RRTs) that grow both globally and locally to sample recently scanned regions. This strategy, often combined with computer vision algorithms has been widely used~\cite{wu19,placed21iros}. The sample-based frontier detector algorithm~\cite{qiao18} reduces the computational load of the previous methods by only storing the nodes of the search tree. 
Frontier identification in 3D maps is less frequent, since 3D maps are more expensive to store and analyze, and are often incomplete due to the sensed volume. Apart from simple search techniques~\cite{dornhege13, dai20}, most methods evaluate map portions incrementally~\cite{zhu15, batinovic21} or along surfaces~\cite{senarathne16}. Alternatively, in~\cite{shen12}, authors propose a method that disperse random particles over the 3D known space. 
No matter the method used, after detecting frontiers, a clustering step is frequently required to prevent the frontier set from being high-dimensional (e.g.,~using K-means~\cite{lu20} or mean-shift~\cite{umari17}).

Shortly after the concept of frontiers was proposed, Newman \textit{et al.}~\cite{newman03} and Stachniss \textit{et al.}~\cite{stachniss04} realized that, for a robot with high uncertainty, \revised{\textbf{potential loop closure}} areas encode more information than frontiers\revised{; the ultimate goal of active SLAM goes beyond simply covering the workspace: to improve the accuracy of localization and mapping}.
Similarly, Grabowski \textit{et al.}~\cite{grabowski03} observe that regions of interest
where sensor readings overlap may be more informative than new frontiers.
In other words, these works explicitly account for the \emph{exploration-exploitation dilemma} in the frontier detection step. It is a common practice in active SLAM to include potential loop closure regions 
---along with frontiers--- in the set of goal candidates~\cite{valencia12,palomeras19b}, or to switch between exploring new frontiers and revisiting known places~\cite{stachniss04, kim15, suresh20}.

In contrast to frontier-based approaches, some active SLAM formulations allow the identification of goal locations locally in the \revised{\textbf{robot's vicinity}}. However, note that decisions will be optimal only locally and a short decision-making horizon may induce wrong behaviors~\cite{feder99, van12}. This strategy is typical in deep reinforcement learning (DRL) approaches~\cite{tai17, chaplot18, placed20}, for which local optimality is alleviated by network memorization. Following the idea that evaluating larger neighborhoods would lead to more robust decisions, in~\cite{valencia12} authors use RRT-based paths to several configurations over the free space as the action set; and in~\cite{indelman15} the entire environment is considered under the umbrella of continuous-domain optimization. 

\section{Stage 2: Utility Computation} \label{S:4}

The second and main stage in \revised{modular} active SLAM approaches focuses on the evaluation of each possible destination,
in order to estimate the effect that executing the set of actions to reach each destination would have. 
Naive utility formulations using just geometric or time-dependent functions often result in non-desirable behaviors~\cite{charrow15, valencia12, placed21iros}, since they do not properly capture the uncertainty in the belief. The \emph{exploration-exploitation dilemma} can be more effectively solved by quantifying the expected uncertainty of the two target random variables: the robot location and the map. 
\revised{Typically, the different objectives (e.g.,~travelling cost, mapping and localization uncertainty) are aggregated into a single utility function, although there are multi-objective approaches in which they are kept separate and Pareto optimal solutions are sought~\cite{amigoni05, chen19rss, soragna19}.}
There is a plethora of metrics and the choice of 
which one to use mainly depends on the selected way to represent the variables of interest.
Metrics based on information theory usually aim at OG maps, while those based on the theory of optimal design are more suitable for Gaussian distributions. We review each choice below.

\subsection{Naive Cost Functions} 

The simplest (and first-broadly-used) metrics are naive geometric functions, such as the Euclidean distance to the goal location~\cite{yamauchi97}, the time required to reach it~\cite{dai20}, or 
the expected size of the are to visit~\cite{gonzalez02, tovar06, umari17}.
In fact, the latter approximates
the map's entropy, which is strongly related to the number of known cells in an OG map~\cite{makarenko02}. 
Since these metrics are computed over Euclidean or temporal spaces, they can be used regardless of the map representation chosen~\cite{gomez19,palomeras19b, batinovic21}.
Stachniss \textit{et al.}~\cite{stachniss05} show that combining distance and information-based functions results in better exploration strategies, and this has since been a common approach~\cite{chen20}.
However, manual tuning to overcome discrepancies between the multiple terms involved is needed~\cite{julia12, umari17}.

\subsection{Information Theory (IT)}

The most common approach to assess utility in active SLAM uses information theory (IT) to quantify the uncertainty in the  joint belief state. Within it, there exist different metrics that allow for such quantification, although all of them build on the same concept: \rmk{entropy}. The notion of entropy was introduced by Shannon~\cite{shannon48} and can be defined as a measure of a variable's uncertainty, randomness, or surprise; this is in fact strongly related to its associated information~\cite{cover99}. 

Early exploration strategies use only the map representation as the variable of interest~\cite{yamauchi97, colares16, dai20}, thereby assuming no error in the robot localization. However, soon after the first of these works emerged, it was observed that high uncertainty in the robot state estimation leads to wrong expected map uncertainties~\cite{bourgault02}. 
The entropy of the SLAM posterior after executing a candidate action can be computed as~\cite{stachniss05}: 
\begin{gather}
    \nonumber \mathcal{H}\left[p(\boldsymbol{x}, \boldsymbol{m}|\boldsymbol{h}, \hat{\boldsymbol{z}}, \boldsymbol{a})\right] \triangleq\\
    \underbrace{%
     \vphantom{ \int_{\boldsymbol{x}} p(\boldsymbol{x}|\boldsymbol{h}, \hat{\boldsymbol{z}}, \boldsymbol{a}) \mathcal{H}\left[p(\boldsymbol{x}|\boldsymbol{x},\boldsymbol{h}, \hat{\boldsymbol{z}}, \boldsymbol{a})\right] d\boldsymbol{x} } 
     \mathcal{H}\left[p(\boldsymbol{x}|\boldsymbol{h}, \hat{\boldsymbol{z}}, \boldsymbol{a})\right]}_{\text{robot's $\mathcal{H}$}} + \underbrace{\int_{\boldsymbol{x}} p(\boldsymbol{x}|\boldsymbol{h}, \hat{\boldsymbol{z}}, \boldsymbol{a}) \mathcal{H}\left[p(\boldsymbol{m}|\boldsymbol{x},\boldsymbol{h}, \hat{\boldsymbol{z}}, \boldsymbol{a})\right] d\boldsymbol{x}}_{\text{expected conditional map's $\mathcal{H}$}}\, ,\label{eq:shannon}
\end{gather}%
\noindent where $\hat{\boldsymbol{z}}$ are the expected (ML) future measurements, which may be estimated using, e.g.,~ray-casting techniques~\cite{carrillo12}. 

The computation of the previous joint entropy is
intractable in general~\cite{stachniss05}. To overcome this, most approaches resort to \rmk{entropy approximations} that first compute utility of the two variables independently, and then combine them heuristically~\cite{bourgault02, stachniss05, vallve14, suresh20}. Let us first consider the case of graph-based SLAM, in which the problem is described using a graph representation where nodes represent the robot poses and edges encode the the constraints between them; see~\cite{thrun05, durrant06, grisetti10,cadena16}.
The joint entropy in~\eqref{eq:shannon} can be approximated by~\cite{valencia12}: 
\begin{equation}
   \mathcal{H}\left[p(\boldsymbol{x}, \boldsymbol{m}|\boldsymbol{h}, \hat{\boldsymbol{z}}, \boldsymbol{a})\right] \approx \mathcal{H}\left[p(\boldsymbol{x}|\boldsymbol{h}, \hat{\boldsymbol{z}}, \boldsymbol{a})\right]
   + \mathcal{H}\left[p(\boldsymbol{m}|\boldsymbol{h},\hat{\boldsymbol{z}}, \boldsymbol{a})\right]\, . \label{eq:h}
\end{equation}%
\noindent The mismatch between the magnitudes of the addends above
is the main drawback of such approximation, calling for the addition of weighting parameters to balance the contributions of the two terms~\cite{blanco08, carlone14jirs}. Carrillo \textit{et al.}~\cite{carrillo18} circumvent this by embedding a metric of the robot's uncertainty in a combined Shannon-R\'enyi utility function; an
approach that also appears in~\cite{popovic20}. On the other hand, the expectation-maximization (EM) algorithm~\cite{wang20} embeds the impact of robot's uncertainty directly in a virtual map.
%
A similar approximation can be done for particle-filter SLAM, which represents the belief over robot trajectories as a set of particles~\cite{thrun05, durrant06, montemerlo02}. 
The 
integral in~\eqref{eq:shannon} will be now approximated by the weighted mean of all possible solutions (i.e.,~particles)~\cite{stachniss05}.

The first term in~\eqref{eq:shannon}
refers to the robot state entropy, which can be computed as a function of the posterior covariance log-determinant, assuming that it is an $\ell$-dimensional Gaussian distribution with covariance $\boldsymbol{\Sigma}_r\in \mathbb{R}^{\ell\times \ell}$,
\begin{align}
    \mathcal{H}\left[p(\boldsymbol{x}|\boldsymbol{h}, \hat{\boldsymbol{z}}, \boldsymbol{a})\right] = \frac{1}{2} \ln \left( (2\pi e )^\ell \det \left(\boldsymbol{\Sigma}_r\right)\right)\, . \label{eq:robotH}
\end{align}%

On the other hand, the second term is the expected map's entropy, and its computation depends on the representation chosen. For instance, in landmark-based maps it can be computed in the same way as the robot's entropy, under the same assumption~\cite{davison05}. For discrete metric maps, and assuming cells independent from each other, it can be defined as~\cite{stachniss09}:
\begin{align}
    \mathcal{H}\left[p(\boldsymbol{m}|\textbf{x},\boldsymbol{h}, \hat{\boldsymbol{z}}, \textbf{a})\right] = - \textstyle\sum_{c\in\boldsymbol{m}} \theta_c \log \theta_c\, , \label{eq:mapH}
\end{align}%
\noindent with $\theta_c=p(c)$ being the occupancy probability of cell $c$. This entropy measure has been used in both 2D~\cite{makarenko02, vallve14} and 3D OG maps~\cite{palomeras19, dai20, lu20}. 
 More efficient approaches that only evaluate of cells in the robot's vicinity have been proposed in the context of particle-filter SLAM~\cite{blanco08, du11, carlone14jirs}.

The most common metric to assess utility in active SLAM is not Shannon's entropy of the SLAM posterior, but its expected reduction.
This utility function is known as \rmk{mutual information} (MI)~\cite{bourgault02, stachniss03} and is defined as the difference between the entropy of the actual state and the expected entropy after executing an action,
i.e.,~the information gain:
\begin{equation}
    \mathcal{I}(\boldsymbol{a}) \triangleq \underbrace{\mathcal{H}\left[  p\left(\boldsymbol{x},\boldsymbol{m}|\boldsymbol{h}\right)\right]}_{\text{current } \mathcal{H}} - \underbrace{\mathbb{E}\left[\mathcal{H}[ p(\boldsymbol{x},\boldsymbol{m}|\boldsymbol{h}, \hat{\boldsymbol{z}}, \boldsymbol{a})]\right]}_{\text{expected }\mathcal{H}\text{ for candidate } \boldsymbol{a}}\, , \label{eq:ig}
\end{equation}%
where expectation is taken w.r.t. $\hat{\boldsymbol{z}}$.

\rmk{Kullback-Leibler divergence} (KLD) or relative entropy~\cite{kullback51} has also been used as utility function. KLD measures the change in the form of a pdf (as MI), but also how much its mean has translated~\cite{mihaylova02}. It is defined as follows:
\begin{align}
     \mathcal{D}_{KL}\left(p_1|p_2\right) &\triangleq \mathbb{E}\left[ \log \frac{p_1(\boldsymbol{x})}{p_2(\boldsymbol{x})} \right] = \sum_{\boldsymbol{x}} p_1(\boldsymbol{x}) \log \frac{p_1(\boldsymbol{x})}{p_2(\boldsymbol{x})},
\end{align}
\noindent with $p_1(\boldsymbol{x})$ and $p_2(\boldsymbol{x})$ the prior and posterior distributions (as in MI)~\cite{houthooft16}, or the estimated and true posteriors assuming the latter can be somehow approximated~\cite{fox03, kontitsis13, carlone14jirs}.

For OG maps, the three metrics above (entropy, MI, and KLD) ultimately rely on counting the number of cells in a map, being thus discrete and ill-suited for optimization techniques. To mitigate this issue, Deng \textit{et al.}~\cite{deng20, deng20b} propose a differentiable \emph{cost-utility} function for both 2D OG and voxel maps that can be used with continuous optimization methods (albeit the approach still assumes perfect robot localization).

In the context of information-theoretic planning, there exists a problem variation in which the uncertainty of only a subset of variables is reduced. The motivation comes from the fact that maximizing information of all variables does not always imply maximizing that of the subset of interest. This problem variation has been referred to as 
\rmk{focused active inference}~\cite{levine13}. In general, focused active inference is more computationally  intensive than the standard case, since it requires marginalization of the (posterior) Fisher information matrix via, e.g.,~Schur complement. Kopitkov and Indelman~\cite{kopitkov17, kopitkov19} present a method 
based on the matrix determinant lemma that does not require the posterior covariance to calculate entropy considering both the unfocused (entropy over all variables) and focused (entropy over a subset of variables) cases.

\subsection{Theory of Optimal Experimental Design (TOED)}

There exists a second group of utility functions built upon optimal design theory (TOED) that tries to quantify uncertainty directly in the task space (i.e.,~from the variance of the variables of interest). Unlike information-theoretic metrics that target binary probabilities in the grid map, task-driven metrics apply to Gaussian variables. Following TOED, a set of actions to execute in active SLAM will be preferred over another if the covariance of the joint posterior is smaller, i.e.,~the posterior covariance matrix, $\boldsymbol{\Sigma}$, has to be minimized. In order to compare matrices associated to different candidates, several functions ---known as \emph{optimality criteria}--- have been proposed, such as the trace (originally known as $A$-optimality)~\cite{chernoff53}, its maximum/minimum eigenvalue ($E$-optimality)~\cite{ehrenfeld55}, or its determinant ($D$-optimality)~\cite{wald43}. 
%
The latter was often disregarded in active SLAM because its traditional formulation did not allow for checking task completion and generated precision errors ($\det(\boldsymbol{\Sigma})\to 0$ rapidly when there are low-variance terms)~\cite{mihaylova02, sim05}. However, Carrillo \textit{et al.}~\cite{carrillo12} show these problems can be solved using Kiefer's formulation of $D$-optimality~\cite{kiefer74}, thus re-establishing the latter as an effective measure of uncertainty for active SLAM.

On the basis of TOED, Kiefer~\cite{kiefer74} proposes a family of mappings $\|\boldsymbol{\Sigma}\|_p:\mathbb{R}^{n\times n}\to\mathbb{R}$, parametrized by a scalar $p$:
\begin{equation}
    \|\boldsymbol{\Sigma}\|_p \triangleq \left( \frac{1}{n}\text{trace}(\boldsymbol{\Sigma}^p) \right) ^ \frac{1}{p}\, , \label{eq:kiefer}
\end{equation}
\noindent which can be particularized for the different values of $p$ and expressed in terms of the eigenvalues of $\boldsymbol{\Sigma}$, $(\lambda_1,\dots,\lambda_\ell)$, by leveraging the properties of the matrix power:
\begin{equation}
    \|\boldsymbol{\Sigma}\|_p =
        \begin{cases}
            \left( \frac{1}{n}\sum\limits_{k=1}^n \lambda_k^p \right)^{\frac{1}{p}}, & \text{if } 0<|p|<\infty\\
            \exp \left(\frac{1}{n} \sum\limits_{k=1}^n \log(\lambda_k) \right), & \text{if } p=0
        \end{cases} \, . \label{eq:utility} 
\end{equation}

In essence, utility functions are functionals of the eigenvalues of $\boldsymbol{\Sigma}$. The boundary cases $p=\{0,\pm \infty\}$ and $p=\pm 1$ result in the four modern optimality criteria:
\begin{itemize}
  \item \textit{T}-optimality criterion ($p=1$) captures the average variance:
  \begin{equation}
    T{\text -} opt \triangleq \frac{1}{n}\textstyle\sum_{k=1}^n \lambda_k\, . \label{eq:topt}
  \end{equation}
  
  \item \textit{D}-optimality criterion ($p=0$) captures the volume of the covariance (hyper) ellipsoid:
  \begin{equation}
    D{\text -} opt \triangleq \exp \left(\frac{1}{n} \textstyle\sum_{k=1}^n \log(\lambda_k) \right)\, . \label{eq:dopt}
  \end{equation}
  
  \item \textit{A}-optimality criterion ($p=-1$) captures the harmonic mean variance, sensitive to a lower-than average value:
  \begin{equation}
    A{\text -} opt \triangleq \left(\frac{1}{n}\textstyle\sum_{k=1}^n \lambda_k^{-1}\right)^{-1}\, . \label{eq:aopt}
  \end{equation}
  
  \item $E$-optimality criterion ($p\to\pm\infty$) captures the radii of the covariance  (hyper) ellipsoid:
  \begin{align}
    E{\text -} opt &\triangleq \min (\lambda_k : k=1,...,n)\, , \label{eq:eoptmin} \\
    \tilde{E}{\text -} opt &\triangleq \max (\lambda_k : k=1,...,n)\, . \label{eq:eoptmax}
  \end{align}
\end{itemize}

Optimality criteria were first used in active SLAM by Feder \textit{et al.}~\cite{feder99}, where utility was computed as the area of the covariance ellipses describing the uncertainty in the
joint posterior. Since then, many active SLAM methods based on TOED have been proposed, mostly based on $T\text{-}opt$~\cite{sim05, leung06} and, recently, $D\text{-}opt$~\cite{placed20, suresh20}. Even so, 
IT-based methods remain the most popular.
Note that both the map and robot 
uncertainties must be described by a covariance matrix $\boldsymbol{\Sigma}\in\mathbb{R}^{n\times n}$, either by using a full covariance matrix in landmark-based representations (i.e.,~$n\gg\ell$) or by including the effect of the map's uncertainty in $\boldsymbol{\Sigma}_r$ (and thus $n=\ell$)~\cite{carlone17}.

\rmk{Monotonicity.} One of the most important assumptions in active SLAM is that uncertainty increases as exploration takes place. However, the seminal work in~\cite{carrillo15} notes how monotonicity is lost for some utility functions under certain conditions, concluding that only $D\text{-}opt$ guarantees this property and is thereby the only appropriate utility function for this task. Kim and Kim~\cite{kim17} and Rodr\'iguez-Ar\'evalo \textit{et al.}~\cite{rodriguez18} demonstrate, however, that rather than on the utility function chosen, monotonicity depends on how the error and uncertainty are represented. In~\cite{rodriguez18}, the authors prove that only differential representations guarantee monotonicity for all utility functions. 
In summary, representation of uncertainty is a key issue in active SLAM, since certain representations do not guarantee its monotonicity property during exploration, and thus may lead to incorrect decisions.

\subsection{The Graphical Structure of the Problem}

Quantification of uncertainty via scalar mappings of the covariance matrix may be a computationally intensive task, mostly due to the fact that the covariance is a large and dense matrix. Therefore, most works resort to reasoning over the Fisher information matrix (FIM), i.e.,~the inverse of the covariance, which is generally sparser. Still, their evaluation is expensive, especially for large state spaces. To circumvent this issue, some works have proved that analyzing the connectivity (i.e.,~Laplacian) of the underlying pose-graph in active graph-SLAM is equivalent to computing optimality criteria. 
\revised{The link between graph and optimum design theories can be traced back to Cheng~\cite{cheng81}, who related the number of spanning trees of concurrence graphs with $D$-optimal incomplete block designs.}
Khosoussi \textit{et al.}~\cite{khosoussi14} 
\revised{show} 
that classical $D\text{-}$ and $E\text{-}opt$ are related to the number of spanning trees of the \revised{SLAM pose-}graph and its algebraic connectivity, respectively, for the case of 2D graph-SLAM with constant uncertainty along the trajectory. 
In~\cite{khosoussi19} and~\cite{chen21}, these results are extended to the $\mathbb{R}^n\times SO(n)$ synchronization problem, and also relate $T\text{-}opt$ to the average node degree of the graph. 
Placed and Castellanos~\cite{placed21iros, placed21} study the general active graph-SLAM problem formulated over the Lie group $SE(n)$; showing the existing relationships between modern optimality criteria of the FIM and connectivity indices when the edges of the pose-graph are weighted appropriately, and reporting substantial reductions in computation time. 
These results have been used in
 coverage problems~\cite{chen20b}, multi-robot exploration~\cite{chen20}, \revised{active visual SLAM~\cite{placed22b},} or to develop a \emph{stopping criterion}~\cite{placed22}. 

The graph structure of the problem has also been recently exploited in conjunction with IT utility functions.
Kitanov and Indelman~\cite{kitanov19} relate the number of spanning trees of the graph to entropy (which ultimately depends on the covariance determinant) and its node degree to Von Neumann entropy. The latter has been also applied to the focused case, thus relating the graph topology to the marginalized FIM~\cite{shienman21}.

\section{Stage 3: Action Selection and Execution} \label{S:5}

Once every possible destination has an associated utility value, the last stage of active SLAM involves the selection of the optimal destination. This can be formulated as an optimization problem w.r.t. the set of actions to reach every possible goal location, cf.~\eqref{eq:argmax_eq}. When the set of candidate destinations is discrete (and typically consists in a handful of options), the solution of the optimization can be obtained via enumeration~\cite{yamauchi97, dornhege13, umari17}. For the case of TOED-based utility functions, it will be a minimization or maximization problem depending on whether the covariance ($\boldsymbol{\Sigma}$) or the FIM ($\boldsymbol{\Phi}$) is analyzed. Since $\boldsymbol{\Sigma} = \boldsymbol{\Phi}^{-1}$ and \revised{$\|\boldsymbol{\Sigma}\|_p = \left(\|\boldsymbol{\Phi}\|_{q}\right)^{-1}\forall p$ with $q=-p$}, the optimization problem is
\begin{align}
    \boldsymbol{a}^\star = \argmin_{\boldsymbol{a}\in\mathcal{A}} \| \boldsymbol{\Sigma}\|_p = \revised{\argmax_{\boldsymbol{a}\in\mathcal{A}} \ \| \boldsymbol{\Phi}\|_{q}}\, ,
\end{align}%
\revised{where $\|\cdot\|_p$ refers to Kiefer's optimality criteria, see~\eqref{eq:kiefer}. }

Information-based utility functions will seek to minimize entropy (or, equivalently, to maximize MI). Following~\cite{valencia12}, the optimal set of discrete actions can be found as:
\begin{align}
    \boldsymbol{a}^\star = \argmax_{\boldsymbol{a}\in\mathcal{A}} \ \mathcal{I}_G = \argmin_{\boldsymbol{a}\in\mathcal{A}} \ \mathcal{H}\left[ p(\boldsymbol{x},\boldsymbol{m}|\boldsymbol{h}, \hat{\boldsymbol{z}}, \boldsymbol{a})\right]\, .
\end{align}%

In any case, after selecting the most informative destination, it all comes down to navigating to it using, e.g.,~sampling-based planning methods as RRT~\cite{lavalle01}, probabilistic road maps (PRM)~\cite{kavraki96}, or their asymptotically optimal variants~\cite{karaman11}. Note that despite selecting the optimal destination among a discrete set of candidates, the executed path to reach it rarely represents an optimal solution for the original problem~\eqref{eq:argmax_eq}; this suboptimality is caused by decoupling
the problem into first computing and evaluating a set of goal locations, and then computing a path to one of these goals.

\section{Belief-space Planning and \\ Continuous-space Optimization} \label{S:6}

As a potential solution to the suboptimality induced by classical decoupled approaches, there exists a second family of methods in which the future trajectory of the robot is directly optimized. These methods represent an alternative solution to the \revised{modular} scheme and may be divided into two categories, depending on whether they discretize the action space or not.
The first category relies on path planning algorithms to generate a discrete set of candidate paths towards the unknown space, in order to later evaluate their utility. Works from Oriolo \textit{et al.}~\cite{oriolo04} and Freda \textit{et al.}~\cite{freda05} are among the first to apply these algorithms for exploration, evaluating robot configurations inside the not-previously-sensed free space. 
In contrast to discrete frontier optimization, that compares utility only at candidate locations, these methods evaluate it over the paths to reach them, guaranteeing that the path to execute is optimal among the considered subset.
\revised{Bonetto \textit{et al.}~\cite{bonetto21, bonetto22} go one step further and optimize exploration in all three steps of modular approaches, considering not only the destination and the path to reach it, but also its execution.}

On the other hand, globally optimal solutions have been considered under the umbrella of continuous-state POMDPs. Despite their resolution would ideally require to compute a policy over the infinite-dimensional space of posteriors of the joint state space~\cite{van12b} and computing an exact solution is known to be intractable in general~\cite{madani99}, active SLAM as a continuous-state POMDP can be approximately solved under the frameworks of 
belief-space planning (BSP) or optimal control. 
Such optimization techniques require a continuous utility function, which can be obtained directly from complex continuous representations of the environment~\cite{jadidi18} or inferred from discretized representations. For example, Vallv\'e and Andrade-Cetto~\cite{vallve14} compute a dense entropy field from the posteriors' evaluation over the discretized configuration space.

\subsection{Belief-Space Planning (BSP)}

Continuous-domain BSP optimizes the future trajectory of the robot without discretizing the action space, but rather performing a continuous optimization. 
Bai \textit{et al.}~\cite{bai14} and Kontitsis \textit{et al.}~\cite{kontitsis13} use sampling-based methods to maximize an objective function that rewards uncertainty reduction and goal achievement. 
Platt \textit{et al.}~\cite{platt10} apply linear quadratic regulation (LQR) to compute locally optimal policies. Van Den Berg \textit{et al.}~\cite{van12} relax the assumption that future observations are consistent with the current robot pose belief (ML observations). Indelman \textit{et al.}~\cite{indelman15} extend~\cite{van12} to the case where the belief describes both robot poses and unknown landmarks in the environment, while also modeling missed observations. Porta \textit{et al.}~\cite{porta05} generalize value iteration to continuous-state POMDPs while assuming state-dependent reward functions. Van den Berg \textit{et al.}~\cite{van12b} present a highly efficient method for solving continuous POMDPs in which beliefs can be modeled using Gaussian distributions over $\mathcal{S}$. Prentice and Roy~\cite{prentice09} develop a belief-space variant of the PRM algorithm called the belief road map (BRM), incorporating predicted uncertainty of future position estimates into the planning process. Valencia \textit{et al.}~\cite{valencia13} contribute a pose-SLAM path-planning approach that leverages the BRM to find a path to the goal with the lowest accumulated pose uncertainty. 

\subsection{Active SLAM as Optimal Control}


Converting a POMDP formulation of active SLAM into an equivalent continuous-space MDP, as discussed in Section~\ref{S:2}, leads to a stochastic optimal control problem in general. Depending on the transition and observation models, noise distribution, and the reward function, the problem may be simplified further. Le Ny \textit{et al.}~\cite{leny09} and Atanasov \textit{et al.}~\cite{atanasov14} show that when the transition and observation models are linear in the state $\boldsymbol{s}$ and the noise is Gaussian, then the time evolution of the belief state $b_t$ may be obtained by the Kalman filter and the covariance is \emph{independent} of the measurement realizations. If the reward function $\rho$ depends only on the covariance, as 
for the MI, active SLAM reduces from a stochastic to a deterministic optimal control problem. Deterministic optimal control problems are easier to solve, and techniques such as linear quadratic Gaussian (LQG) regulation~\cite{koga21} or search-based~\cite{atanasov14,schlotfeldt19}, and sampling-based~\cite{hollinger14,lan16,kantaros19} motion planning are applicable. If the assumptions necessary for the deterministic reduction cannot be satisfied, the stochastic active SLAM problem may be solved by obtaining an open-loop control sequence under deterministic dynamics first, followed by a closed-loop feedback policy, under stochastic dynamics linearized around open-loop trajectory~\cite{koga22}.

In the presence of state or action constraints, the optimal control formulation of active SLAM can be approached using differential dynamic programming (DDP) or model predictive control (MPC). Rahman and Waslander~\cite{rahman21} introduce an augmented Lagrangian formulation of iterative LQG, which captures belief-state constraints via a penalty function. The approach iterates between iLQG trajectory optimization in an unconstrained stochastic optimal control problem and Lagrange multiplier updates for the penalty function. This and several other works~\cite{murali19,rahman21,koga21} develop differentiable formulations of sensor field-of-view constraints amenable to gradient-based optimization. 
Carlone and Lyons~\cite{Carlone14icra} split the environment into convex regions and formulate the problem using mixed-integer programming.
Chen \textit{et al.}~\cite{chen20b} employ a spectrahedral description of the convex hull of the space of orientations and relax non-convex obstacle constraints using a convex half-space representation. 

Striking a suitable balance between exploration and exploitation in active SLAM is challenging because the effects of potential future loop closures are not easy to capture in the predicted evolution of the belief $b_t(\boldsymbol{s}_t)$. Leung \textit{et al.}~\cite{leung06,leung08} introduce attractor states to guide the robot based on three modes (explore, improve map, and improve localization), determined using uncertainty thresholds. Attractor states were combined with a right-invariant extended Kalman filter in \cite{xu21} to achieve active range-bearing landmark-based SLAM.

\section{Deep-Learning-based Approaches} \label{S:7}

Advances in DL have created new opportunities in using neural networks to solve active SLAM; 
these techniques follow a completely different scheme, circumventing the split into three stages that characterizes \revised{modular} approaches. Usually, goal identification is not required due to the chosen action set, and utility computation and selection of the best action are both embedded in the network. In this section, we particularly focus on DRL methods for autonomous robotic exploration and discuss the design of the state, action, and reward spaces, as well as the problems of partial observability, generalization, and the necessity for training environments.

\subsection{Deep Reinforcement Learning (DRL)}

A question that arose in the early work on learning-based active SLAM was which type of learning was suitable for this decision-making problem, in which (i) agents must directly learn from interaction with the environment, (ii) states may not be fully observable, and (iii) policies have to generalize to other scenarios in which \emph{a priori} knowledge is nonexistent. 
\revised{This premise soon led the community to explore DRL, building on existing methods that approached active SLAM with RL~\cite{martinez09} and using neural networks to represent the policies or value functions. Within DRL, model-free techniques have been the center of attention, although isolated approaches that combine them with model-based learning do exist~\cite{karkus17}. Methods based on supervised learning can also be found in the literature~\cite{bai17, chen19b}, although they are a minority.}
Contrary to \revised{model-based active SLAM}, the computational effort in DRL approaches is mostly confined to the training phase, while the testing phase reduces to a forward pass on the network. \revised{However, the behavior depends entirely on the model learned from training data, thus limiting its generalization to novel operational conditions.}

The great success of the work from Mnih \textit{et al.}~\cite{mnih15} boosted the research in model-free DRL and several value- and policy-based methods emerged shortly after. The behavior of deep Q-networks~\cite{mnih15} improves using the double~\cite{hasselt16} and double-dueling~\cite{hessel18} architectures. 
Actor-critic techniques combine both value-iteration and policy gradient methods, e.g.,~deep deterministic policy gradient~\cite{lillicrap15}, asynchronous advantage actor-critic~\cite{mnih16}.
See~\cite{arulkumaran17} and~\cite{zeng20} 
for a survey on the methods. Although these strategies were initially proposed for different decision-making problems (e.g.,~video-games), they have been applied to robotic exploration. 


\subsection{On the Reward Function Design and the Action Set}

Tai and Liu~\cite{tai16} are among the first to employ DRL for robotic exploration \revised{in simulation environments}, extracting the next best actions to execute from \revised{raw observations} using a 2-layer Q-network. 
\revised{Convergence to policies valid in more complex and previously unseen scenarios is achieved in~\cite{mirowski17,tai17} with parallel architectures.}
In any case, the above works use purely extrinsic reward functions \revised{(i.e.,~by instrumenting the environment), which ultimately addresses the obstacle avoidance problem rather than active SLAM~\cite{placed20}.}
As a response, the notions of motivation and curiosity~\cite{ryan00} were exploited to design intrinsic rewards, giving origin to \textit{curiosity-driven} methods that motivate agents to visit unknown configurations~\cite{bellemare16}. 
\revised{Chen \textit{et al.}~\cite{chen19learning} and Chaplot \textit{et al.}~\cite{chaplot20} propose holistic, open-source approaches that employ a coverage reward to explore complex 3D simulation environments. The detailed study in \cite{chen19learning} shows the benefits of pre-training and combining inputs from different sources.}
Similarly, the idea of uncertainty minimization led to \textit{uncertainty-aware} approaches. This is the case of~\cite{zhelo18} that encourage the visit of high-covariance states, and~\cite{chaplot18, gottipati19} where the reward encodes the belief accuracy. All these methods \revised{are publicly available except active target localization. Many of the DRL-based methods, including all of the above, aim to directly generate optimal control commands, either discrete~\cite{chen19learning} or not \cite{hu20}. They represent end-to-end solutions in which the safe navigation task is embedded into the network and therefore do not require planning and the SLAM estimates.}

True uncertainty metrics inherited from classic theories have also been introduced in the reward function design, \revised{seeking more robust foundations.
The robot's $D\text{-}opt$ is incorporated in~\cite{placed20} and $T\text{-}opt$ of virtual landmarks in~\cite{chen20c}, whereas the map's MI is used in~\cite{niroui19,li20}. 
Agents trained under this new perspective perform active SLAM in complex scenes, albeit only targeting location or mapping uncertainties. Designing effective reward functions that account for both is still an open problem.
In addition, this new family of methods has promoted the use of learning as a part of the solution rather than a replacement to well-established planning algorithms. 
Utilizing planning and learning together, may make policies easier to learn, generalize better and transfer across platforms. In this vein, Niroui \textit{et al.}~\cite{niroui19} and Chen \textit{et al.}~\cite{chen20c} employ DRL to extract the best candidate among previously-detected frontiers, thereby creating a link with modular approaches.
Li \textit{et al.}~\cite{li20} and Lodel \textit{et al.}~\cite{lodel22} use nearby sampled locations instead, but they also leave the motion planning task out of the scope of learning. Chaplot \textit{et al.}~\cite{chaplot20} use different policies to infer long-term (i.e..,~frontiers) and short-term (i.e.,~control commands) goals, linked through a model-based trajectory planner.}

\subsection{Partial Observability and Generalization}

Partial observability and generalization are 
two inherent and often-forgotten concepts in active SLAM. First of all, the uncertainty about the observations and actions taken, and the limited observations make the problem not fully observable. Consequently, agents are unable to distinguish their own true state based on single observations, and learned policies are bound to be suboptimal~\cite{zhu21}. 
\revised{Mirowski \textit{et al.}~\cite{mirowski17} alleviate this by expanding the network inputs with previous observations and rewards.
Hausknecht and Stone~\cite{hausknecht15} demonstrate that recurrent architectures can also handle partial observability, teaching agents to learn about previous data on their own. Long short-term memory units are used for robotic exploration in~\cite{niroui19,mirowski17,chen19learning}, and Karkus \textit{et al.}~\cite{karkus17} embed the computation structure of the belief (and thus the history) in a recurrent neural network.}

The second element intrinsic to active SLAM is the lack of prior knowledge of the environment. \revised{Learning policies that generalize to unseen scenarios is therefore crucial, and currently represents a key limiting factor for learning-based methods.} 
\revised{Overfitting can be mitigated by expanding the sample space (e.g.,~using random starting locations~\cite{zhelo18, niroui19}, considering noise in the observations~\cite{hu21}) or by using sparser network inputs~\cite{zhu21}.
For example, agents trained in~\cite{tai17, yokoyama20} learn policies generalizable to real environments after reducing sensory data to a sparse range input. 
Similarly, Shi \textit{et al.}~\cite{shi19} specifically use sparse range measurements to reduce the simulation-to-reality (\emph{sim-2-real}) gap.
Lodel \textit{et al.}~\cite{lodel22} improve generalization by feeding the network with egocentric limited observations, following~\cite{chen19learning}. 
Chen \textit{et al.}~\cite{chen20c} leverage graph neural networks, in which the inputs are already compressed representations. The task of transferring trained agents to real scenarios is still an open research problem, and few efforts have been made in this direction~\cite{tai17, shi19, li20}}.

\subsection{Training Environments}

The use of DRL introduced a major challenge during training: the need of a simulation environment to acquire data online. Unlike supervised methods, training with offline data is not possible and real-world training seems infeasible. To overcome this problem, some works use their own simplified simulation scenarios, \revised{thus limiting the network inputs to ground-truth data or range perfect observations. To use more realistic data that bridge the gap from simulation to physical robots, more complex simulators need to be used in training.}

\emph{Stage}~\cite{gerkey03} is one of the simplest engines used in the literature~\cite{niroui19}, although it restricts perception to two-dimensional bitmapped environments. \emph{Gazebo}~\cite{koenig04} is a much more complete simulator which allows for 3D simulations, realistic rendering, visual sensors, 
etc. In addition, it is tightly integrated into the widespread Robotic Operating System (ROS), which makes its use commonplace~\cite{tai16,tai17,placed20}.
\emph{CoppeliaSim/V-REP}~\cite{rohmer13} also allows for online mesh manipulation, but it is not an open-source solution and is less integrated into ROS, limiting its adoption. 
Combination of a physics engine (i.e.,~robot motion and sensor models) with a DRL framework is not always straightforward. Zamora \textit{et al.}~\cite{zamora16} present a powerful framework by integrating the RL toolkit OpenAI Gym~\cite{brockman16} with ROS and Gazebo.

In contrast to the above platforms, initially designed for robotics and later adapted to DRL, there is a second family of simulators born in the age of AI. They tend to prioritize training speed over the breadth of simulation capabilities. \emph{DeepMind Lab}~\cite{beattie16} allows agents to move discretely in low-textured, game-like scenarios, and provides access to a visual sensor and velocity.
\emph{Habitat-Sim}~\cite{savva19} takes a  leap forward by supporting physics simulation and different robot and visual sensor models. More interestingly, it has the powerful capability of rendering simulation environments from image datasets, e.g.,~Replica~\cite{straub19}.
\emph{iGibson}~\cite{li22} also provides fast visual rendering and physics simulation, and includes simulation of lidar and optical flow sensors. 
The ROS ecosystem is already integrated in~\cite{li22}, whereas~\cite{savva19} requires the use of external libraries. Despite their potential, none of these platforms has yet been used for 
DRL in the context of 
active SLAM.

\section{Multi-robot Active SLAM}\label{S:8}

The active SLAM problem can be extended to a multi-agent setting, where $n$ robots optimize their sensing trajectories collaboratively to estimate a common map $\boldsymbol{m}\in\mathcal{M}$ of the environment. Each robot has its own state space $\mathcal{X}_i$ and action space $\mathcal{A}_i$. Applying an active SLAM algorithm to the joint state space $\mathcal{S} = \mathcal{X}_1\times \cdots \times \mathcal{X}_n \times \mathcal{M}$ and joint action space $\mathcal{A} = \mathcal{A}_1\times \cdots \times \mathcal{A}_n$ can generate desirable behavior but becomes computationally infeasible as the number of robots increases because the complexity of centralized algorithms scales exponentially with $n$~\cite{atanasov15}. Such algorithms also require collecting all robot measurements and performing joint optimization at a centralized server before communicating the planned actions back to the individual robots. If the robot team is small and connectivity is maintained at all times, centralized algorithms can be used to plan all robot trajectories simultaneously. For example, Charrow \textit{et al.}~\cite{Charrow_AuRo14} achieve multi-robot target tracking by maximizing the MI between the target location and range-only observations over a set of motion primitives. However, larger teams with intermittent communication and limited onboard computation require decentralized algorithms, where individual robots solve smaller instances of the active SLAM problem, or fully distributed algorithms, where the robots exchange information only with their neighbors.
Kantaros \textit{et al.}~\cite{kantaros19,kantaros21} propose an informative planning\footnote{
Informative path planning can be considered a generalization of active SLAM to include objectives beyond the quality of localization and mapping, e.g.,~for target tracking or environmental monitoring.} technique which constructs random trees of control sequences and is particularly simple to distribute. The algorithm scales to very large numbers of sensors and targets and is probabilistically complete and asymptotically optimal.

A particularly important instance of the problem is \rmk{collaborative multi-robot exploration}, where the robots aim to coordinate how to efficiently explore different regions of the environment. Early works such as~\cite{burgard00, burgard05} present an approach for choosing appropriate frontiers,
while simultaneously taking into account their utility and the cost of reaching them. Each time a target point is assigned to some specific robot,  the utility of the unexplored area visible from that frontier is reduced. This mechanism is used to assign different frontiers to different robots. Colares and Chaimowicz~\cite{colares16} develop a decentralized multi-robot formulation of the classical frontier-based exploration method. The authors use an objective function that captures the frontier entropy and distance, and a robot coordination factor that penalizes regions that other robots are already exploring. 
Atanasov \textit{et al.}~\cite{atanasov15} consider a multi-agent active information acquisition problem, in which an information measure is maximized over a discrete space of agent trajectories, and propose a decentralized planning scheme using coordinate descent in the space of agent trajectories. Schlotfeldt \textit{et al.}~\cite{schlotfeldt18} introduce an anytime search-based planning formulation that progressively reduces the suboptimality of the multi-agent plans while respecting real-time constraints. 
Instead of using search-based planning, Ossenkopf \textit{et al.}~\cite{ossenkopf19} generate candidate robot actions using RRT*. The sampling is biased to prioritize exploration, map improvement, or localization improvement. The map and robot state entropy is evaluated along the planned trajectories in two stages: short-horizon exact computation using filter updates, and long-horizon approximation using predicted loop closures. 
Lauri \textit{et al.}~\cite{lauri17} introduce a decentralized $\rho$-POMDP,
allowing the specification of an information-theoretic objective. The authors show that a multi-agent A* algorithm that searches the joint policy space can be applied to belief-dependent rewards to achieve cooperative target tracking with periodic communication.
Hu \textit{et al.}~\cite{hu20} design a hierarchical control approach for cooperative exploration, combining a high-level region-assignment layer and a low-level safe-navigation layer. The former uses dynamic Voronoi partitions to assign different regions to individual robots; 
the latter achieves collision-free navigation to successive frontier points using DRL.

Another important instance is \rmk{collaborative multi-robot active estimation}, where the goal is to seek actions that actively reduce the uncertainty over  relevant random variables. 
For instance, Kontitsis \textit{et al.}~\cite{kontitsis13} develop a multi-robot active SLAM method that uses a relative entropy optimization technique~\cite{botev13} to select trajectories which minimize both localization and map uncertainties.
Indelman~\cite{indelman18} develops a collaborative multi-robot BSP framework, which incorporates reasoning about future observations of environments that are unknown at planning time. That approach has been extended in~\cite{regev16} to a decentralized setting. 
Best \textit{et al.}~\cite{best16} propose the \emph{self-organizing map} algorithm, considering the problem of multi-robot path planning for active perception and data collection tasks. 
Chen \textit{et al.}~\cite{chen20} leverage graph connectivity indices and their relationship to optimality criteria to achieve multi-robot active graph-SLAM. Each robot aims to improve the pose graphs of the other agents by sharing its observations when it moves near areas where they have low connectivity.

\section{Open Research Questions} \label{S:9}

Active SLAM still requires much research in order to 
support the deployment of fully autonomous robots in complex environments. Many are the challenges and research fields involved, so cooperation between them is crucial to achieve real-world impact. In this section, we present some of what we consider the most important research questions. 
Although some of them are long-known challenges and are already under intense investigation, others have not received such attention.

\subsection{Prediction Beyond Line-of-sight}

Resolution of active SLAM relies on fast and precise predictions of future states for the variables of interest. The accurate prediction of the scene and robot pose after executing a set of candidate actions can be the difference between making the right decision or not. The expected sensed space and the resulting map representation have traditionally been predicted using a sensor model together with ray-casting techniques~\cite{burgard05, carrillo18}. Recent related work, however, addresses the problem of scene completion and occupancy anticipation from a DL perspective. Fehr \textit{et al.}~\cite{fehr19} use a neural network to augment the measurements of a depth sensor and Ramakrishnan \textit{et al.}~\cite{ramakrishnan20} directly predict augmented OG maps beyond the sensor's field-of-view using auto-encoders (AE). Rather than using raw sensor measurements, Katyal \textit{et al.}
~\cite{katyal21} and Hayoun \textit{et al.}~\cite{hayoun20} extend an input OG map beyond the line-of-sight also using AE. Shrestha \textit{et al.}~\cite{shrestha19} predict maps of occupancy probabilities instead with variational AE. Dai \textit{et al.}~\cite{dai18} perform scene prediction over 3D SDF-based maps. All these methods seem promising for fast and precise online map prediction beyond line-of-sight, although their integration into active SLAM is yet to be done and brings with it numerous challenges. How does scene prediction behave in unstructured environments? How to account for uncertainty? Is measurement prediction more reliable and informative than map prediction? How to predict the effect of only a certain set of non-myopic actions in the map rather than augmenting the whole scene? Regarding the latter,~\cite{richter17, richter18} and~\cite{asraf20} subordinate predictions to candidate actions.

On the other hand, the robot state is straightforwardly estimated using motion models or path planners. However, the prediction of its associated uncertainty is not trivial and requires more attention. Work from Asraf and Indelman~\cite{asraf20} is among the very few efforts made to combine data-driven scene prediction with BSP. In addition, they use the predicted observations to forecast the posterior uncertainty over the robot trajectory. Besides the robot's movement, it is the appearance of loop closures (exploitation) that significantly affects the new states' uncertainty, thus making its forecast critical. Despite some isolated works have partially studied this problem~~\cite{stachniss04, fairfield10},
it still remains as an open challenge.

\subsection{From Active SLAM to Active Spatial Perception}

Most active SLAM approaches reason over geometric representations of the environment (e.g.,~OG maps). 
However, when we explore new environments as humans, we are mostly interested in semantic elements of the environment (e.g.,~presence of objects, rooms) rather than the shape of the environment \emph{per se}.
Modern SLAM systems are now able to build 3D metric-semantic maps
in real-time from semantically labeled images, see~\cite{rosinol20} and the references therein. 
These maps include both occupancy information and semantic labels of entities (e.g.,~chairs, tables, humans, etc.) in the environment.
Very recent work goes even further and develops spatial perception systems that infer hierarchical map representations, in the form of 3D scene graphs~\cite{rosinol21,armeni19,hughes22}. They symbolize high-level representations of an environment, that capture from low-level geometry (e.g.,~a 3D mesh of the environment) to high-level semantics (e.g.,~objects, people, rooms, buildings, etc.).   
While there is a growing amount of work in estimating these high-level representations from sensor data, their use in active SLAM is still uncharted territory. 
Very early effort in this direction includes the work from
Ravichandran \emph{et al.}~\cite{ravichandran22}, which focuses on object search using 3D scene graphs.

Active metric-semantic information acquisition, or \emph{Active Spatial Perception}, has the potential to impact many aspects of robot autonomy: 
(i) by leveraging semantic knowledge,  a robot can more effectively predict unseen space (e.g.,~predict the presence of rooms or objects in each room),
(ii) the use of semantics can further enhance SLAM performance (e.g.,~allowing for novel loop closure detection methods~\cite{hughes22}), 
and (iii) hierarchical representations may enable novel and more computationally efficient planning methods.
However, each opportunity comes with many open research questions, for instance: How to quantify uncertainty over metric-semantic or even hierarchical scene representations? How to leverage hierarchical structures to improve computation? How to perform spatial prediction in hierarchical representations?

\subsection{Robust Online Belief Space Planning and Active SLAM}

Another key aspect is data association, i.e.,~association between measurements and the corresponding  landmarks (or other entities in the map representation). In perceptually aliased and ambiguous environments, determining the correct data association is challenging, and incorrect associations may lead to catastrophic failures. The research community has been investigating approaches for robust perception to allow reliable and efficient operation in ambiguous environments (see, e.g.~\cite{sunderhauf12, olson13, Yang20ral-GNC, indelman16csm, hsiao19, shelly22}). Yet, these approaches focus on inference (rather than planning), i.e.,~actions are given. Only recently, ambiguous data association was considered also within BSP and, in particular, active SLAM. Pathak \textit{et al.}~\cite{pathak18} incorporate, for the first time, reasoning about future data association hypotheses within a BSP framework, enabling autonomous hypotheses disambiguation.  Another related work in this context is~\cite{hsiao20}, that also reasons about ambiguous data association in future beliefs while utilizing the graphical model presented in~\cite{hsiao19}. A first-moment approximation to Bayesian inference with random sets of targets, known as the probability hypothesis density (PHD) filter, has been successfully applied to active target tracking problems~\cite{dames20}.
However, explicitly considering all possible data associations leads to an exponential growth of the number of hypotheses, and determining the optimal action sequence quickly becomes intractable. 
Shienman and Indelman~\cite{shienman22} recently presented an approach that utilizes only a distilled subset of hypotheses to solve BSP problems while reasoning about data association and providing performance guarantees considering a myopic setting. Nevertheless, BSP and active SLAM in these challenging settings remain open problems. More generally, finding an appropriate simplification of the original BSP or active SLAM problem, which is easier to solve, with no, or bounded, loss in performance, is an exciting and novel direction~\cite{indelman16ral, elimelech22, shienman22, barenboim22ijcai}. 

\subsection{Reasoning in Dynamic and Deformable Scenes}

One of the most common assumptions in active SLAM is to consider the environment static ---or slightly dynamic, at best. Real scenes, however, contain moving agents most of the times, and even deformable elements (e.g.,~clothes, water). Handling these elements would greatly impact the robot's autonomy, its reasoning ability and awareness, and would facilitate its deployment in real-world scenarios.

The study of dynamic environments has long been a topic of interest for the path planning~\cite{van05} and the SLAM~\cite{saputra18} communities; but its investigation in the context of active SLAM has been typically restricted to the action execution step (i.e.,~re-planning)~\cite{trivun15, maurovic17}. However, many other aspects emerge when reasoning with dynamic elements: How to foresee their effects in planning? How to integrate them in the utility function? 
How to maintain a lightweight representation?

Non-rigid environments present an even greater challenge. Planning for mobile robots in deformable environments started receiving some attention a couple of decades ago~\cite{anshelevich00, rodriguez06}.
Medical applications have also stimulated progress on SLAM in deformable environments~\cite{newcombe15, lamarca20}. 
However, to date, no efforts have been made towards developing a deformable active SLAM framework.
We believe this is partly due to the unavailability and complexity of simulators for mobile robots in deformable environments, and partly due to the difficulty in extending the current map representations to deformable scenes. 
Given the importance of obtaining accurate robot trajectory estimates towards mapping deformable environments, 
active SLAM can play a major role in this area.

\subsection{Towards Meaningful and Autonomous Stopping Criteria}\label{S:9:d}

Unlike with coverage and exploration in known environments, determining the moment in which the task of active SLAM 
is complete is non-trivial. 
Performing active SLAM is known to be a computationally expensive process: a vast amount of resources is required to estimate and optimize utility online, thereby conditioning the execution of other tasks; therefore, it is crucial to understand when such process can be considered complete and other tasks can be prioritized. 
Cadena \textit{et al.}~\cite{cadena16} already identified this issue as an open research question, but little research has been done on the topic. Even recent active SLAM works still rely on traditional temporal~\cite{carrillo18,placed21iros} or spatial~\cite{chen20b, xu22} constraints to decide when exploration has terminated. These metrics, however, cannot be used in truly unknown environments nor do they assert task completion (see~\cite{placed22}). The use of TOED-based metrics has been identified as a promising tool~\cite{cadena16,lluvia21,placed22} to determine when a given exploration strategy is no longer adding information.
Nevertheless, many questions remain to be answered: How to guarantee task completion? How to transition between exploration strategies? Also, the advent of DRL approaches 
raises a new question: when to stop training?

\subsection{Reproducible Research in Active SLAM}\label{S:9:e}

The increasing attention towards active SLAM creates the need for new benchmarks to objectively evaluate new findings against existing research. 
This has long been a challenge in the robotics community~\cite{del2006benchmarks}, where real-life robotics experiments are often difficult to replicate across research groups. 
In related problems, such as SLAM, static datasets are commonly used for benchmarking.
However, in active SLAM, the agent must interact with the environment, 
making the use of datasets impractical. 
In recent years, a significant effort has been made in robotics to address challenges in benchmarking~\cite{calli2015benchmarking} and reproducibility~\cite{pineau2021improving}.
Despite these efforts, such benchmarks are still lacking in active perception.

Typically, in active SLAM, researchers select a set of scenarios (e.g.,~platform, task, and environment) representative of the desired application, and experiments are conducted in simulation via customized simulators or in the real world via specialized hardware.
While such an evaluation is adequate for validation, the specified scenario may not be general enough or sufficiently specified to be reproduced.
Consequently, one-to-one comparisons are rarely made between approaches.
While targeting more general embodied agents, several open-source datasets~\cite{ammirato2018active} and simulators~\cite{savva19, hall2022benchbot, xu22} show promise for active SLAM research. 
Also, \revised{open-source frameworks (see Table~\ref{tab:works})}
make the comparison and testing of new algorithms straightforward, only by modifying the decision-making portion.
While some works take advantage of these simulators and datasets~\cite{chaplot20}, 
establishing a proper methodology for evaluating active SLAM when it comes to generalization from simulation to the real world remains an open question.
Besides, there is a dire need to establish adequate performance metrics for active SLAM that go beyond commonly-used  exploration time and coverage. Improving the quality of estimates is the main objective of active SLAM, and should therefore be measured. 

\subsection{Practical Applications}

Although active SLAM methods have practical relevance in many real-world problems such as search and rescue, where constructing a sound representation of the environment is time critical, 
 very few practical implementations and deployments of active SLAM have been described in the literature. 
Walter \textit{et al.}~\cite{walter08} propose a partially autonomous system for
underwater ship hull inspection. 
Kim and Eustice~\cite{kim15} deploy a complete active SLAM system. 
Palomeras \textit{et al.}~\cite{palomeras19, palomeras19b} report the autonomous exploration of complex underwater environments for environmental preservation purposes.
Fairfield and Wettergreen~\cite{fairfield10} investigate terrestrial applications and autonomous mapping of abandoned underground mines. 
A roughly similar application but in the archaeological context of catacomb exploration is presented in~\cite{serafin16}.
Strader \textit{et al.}~\cite{strader20} report experiments of active perception in a Mars-analogue environment. 
Finally, assistive mapping examples for office-like environments can be found in~\cite{newman03, lipeng20, hsiao20}.
Aerial applications of active SLAM are significantly less common. Chen \textit{et al.}~\cite{chen20b} propose an MPC framework to address coverage tasks while maintaining low uncertainty estimates.

Overall, there are very few reports of field experiments involving active SLAM systems. Besides, by 2022, there is a large imbalance between the patents using the terms SLAM and active SLAM\footnote{We used ``simultaneous localization and mapping" after:priority:19920101, and "active slam" OR "active simultaneous localization and mapping" after:priority:19920101 as queries search in the Google patents search platform.}, about 39,000 for the former and 31 for the latter. 
This indicates 
that the technology readiness level of active SLAM is not in a deployment phase but in early development. 
Furthermore, it raises the question of whether active SLAM is important for all applications or whether human supervision is still preferred.
Among the roadblocks preventing the transition from theory to applications (including the challenges mentioned in the previous sections), we also remark that the high computational complexity of active SLAM often clashes with application constraints, e.g.,~the low computational budget available on aerial robots.

\section{Conclusions} \label{S:10}

The active SLAM problem, which consists in actively controlling a robot such that it can estimate the most accurate and complete model of the environment, has been a topic of interest in the robotics community for more than three decades, and is now receiving renewed attention ---also thanks to the novel opportunities offered by learning-based methods. 
Despite the role of active SLAM in many applications, 
the disparity and lack of unification in the literature has prevented the research community from providing a cohesive framework, bringing algorithms to maturity, and transitioning them to real applications. In this paper, we take a step toward this goal by taking a fresh look at the problem and creating a complete survey to serve as a guide for researchers and practitioners.

In particular, we present a unified active SLAM formulation under the umbrella of POMDPs, highlighting the most common assumptions in the literature. Then, we discuss the \revised{modular} resolution scheme, which decouples the problem into goal identification, utility computation, and action selection. We delve into each stage, reviewing the most important theories and presenting state-of-the-art techniques. 
We then review alternative approaches that have drawn great interest and have undergone major advances in recent years, including (continuous) BSP and \revised{learning-based approaches}. 
Finally, we discuss relevant work in multi-robot active SLAM. 

Besides discussing the historical evolution and current trends in active SLAM, we also identify the most relevant open challenges in this field. 
These include prediction beyond line-of-sight and active spatial perception, among others.
We also emphasize the need for 
a unified formulation and evaluation metrics that allow for direct comparison between works. Reproducibility and benchmarking need to be addressed for this field to mature and achieve real-world impact. 


\bibliographystyle{IEEEtran}
\bibliography{bibliography}






\end{document}